\documentclass[10pt,twocolumn,letterpaper]{article}

\usepackage[]{cvpr}      
\usepackage[table,xcdraw]{xcolor}
\usepackage{colortbl}

\definecolor{cvprblue}{rgb}{0.21,0.49,0.74}
\usepackage[pagebackref,breaklinks,colorlinks,citecolor=cvprblue]{hyperref}
\definecolor{tabfirst}{rgb}{1, 0.7, 0.7} 
\definecolor{tabsecond}{rgb}{1, 0.85, 0.7} 
\definecolor{tabthird}{rgb}{1, 1, 0.7} 

\title{MaterialFusion: Enhancing Inverse Rendering with Material Diffusion Priors}

\author{
Yehonathan Litman$^1$
\and
Or Patashnik$^2$
\and
Kangle Deng$^1$
\and
Aviral Agrawal$^1$
\and
Rushikesh Zawar$^1$
\and
Fernando De la Torre$^1$
\and
Shubham Tulsiani$^1$
\and
$^1$Carnegie Mellon University
\and
$^2$Tel Aviv University
\and
\url{https://yehonathanlitman.github.io/material_fusion}
}

\begin{document}
\twocolumn[{%
    \renewcommand\twocolumn[1][]{#1}%
    \maketitle
\vspace*{-1cm}
    \begin{center}
        \centering
        \captionsetup{type=figure}\includegraphics[width=0.87\textwidth]{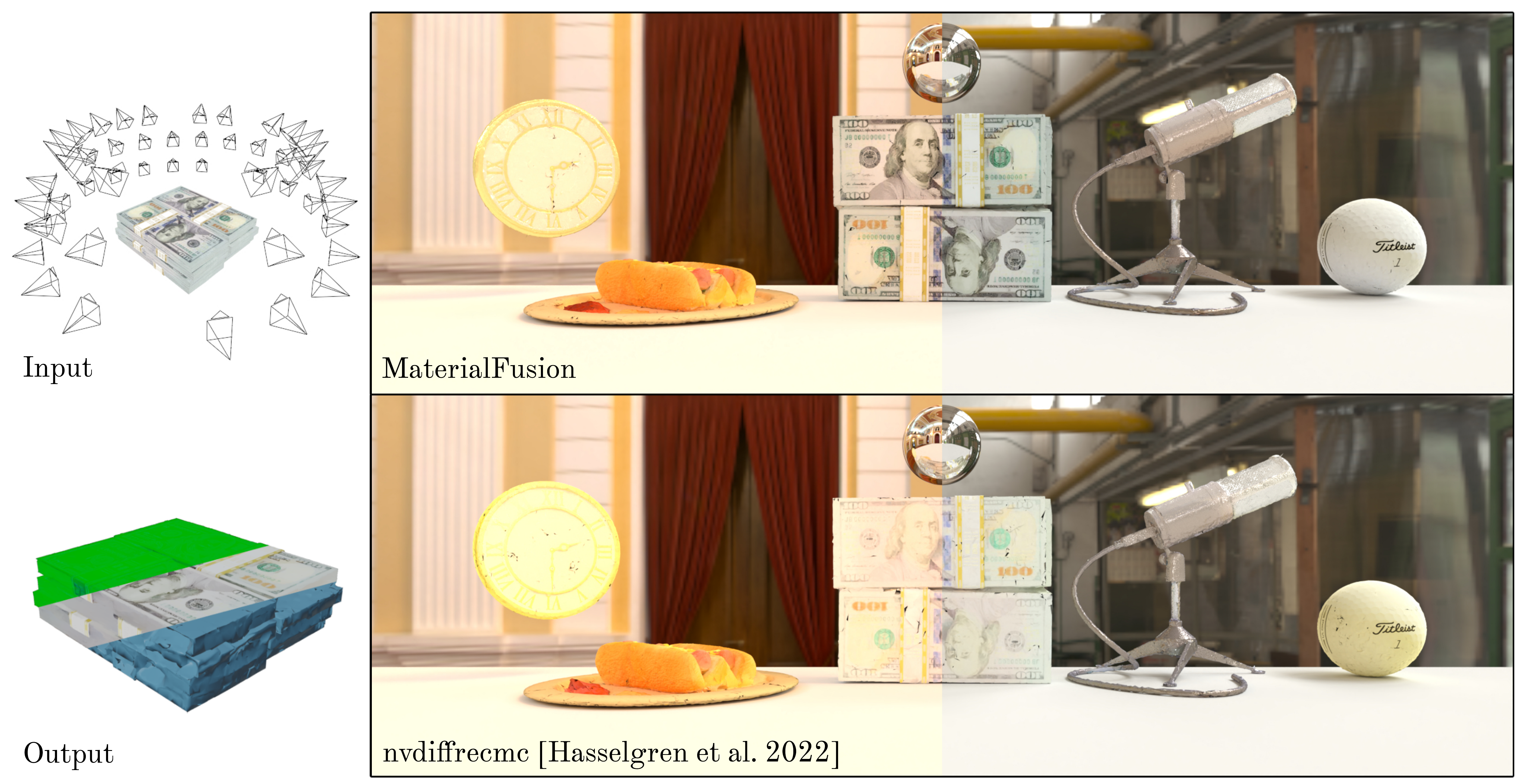}
        \captionof{figure}{
        Given an image set of an object under unknown illumination, MaterialFusion recovers the object's geometry, BRDF appearance, and the environmental illumination, via inverse rendering. Our method utilizes a 2D material diffusion prior to accurately reconstruct these properties. On the left, we display the input image set of the bills alongside the output of the reconstructed properties, visualized as the materials, albedo, and mesh from top to bottom, respectively. On the right, we show different objects rendered under novel lighting conditions with the reconstructed physical properties.}
        \label{fig:teaser}
    \end{center}%
}]

\begin{abstract}

Recent works in inverse rendering have shown promise in using multi-view images of an object to recover shape, albedo, and materials. However, the recovered components often fail to render accurately under new lighting conditions due to the intrinsic challenge of disentangling albedo and material properties from input images. To address this challenge, we introduce MaterialFusion, an enhanced conventional 3D inverse rendering pipeline that incorporates a 2D prior on texture and material properties. We present StableMaterial, a 2D diffusion model prior that refines multi-lit data to estimate the most likely albedo and material from given input appearances. This model is trained on albedo, material, and relit image data derived from a curated dataset of approximately \raisebox{0.5ex}{\texttildelow}12K artist-designed synthetic Blender objects called BlenderVault. We incorporate this diffusion prior with an inverse rendering framework where we use score distillation sampling (SDS) to guide the optimization of the albedo and materials, improving relighting performance in comparison with previous work. We validate MaterialFusion's relighting performance on 4 datasets of synthetic and real objects under diverse illumination conditions, showing our diffusion-aided approach significantly improves the appearance of reconstructed objects under novel lighting conditions. We intend to publicly release our BlenderVault dataset to support further research in this field.

\end{abstract}     
\section{Introduction}

Recently, there has been an increased interest in methods that try to recover 3D representations from 2D images. Novel view synthesis approaches, particularly Neural Radiance Fields (NeRF)~\cite{mildenhall2020nerf} and follow-up works have proven highly effective for accurately representing 3D scenes from posed 2D images. Nevertheless, one of the main drawbacks of these approaches is relighting, since novel view synthesis methods bake in all the lighting information into the 3D representation, rather than disentangling it from the underlying scene data. In this paper, our goal is to infer relightable 3D representations that factorize these properties, allowing for the editing of materials, geometry, and lighting independently.

Some approaches do pursue factorized, relightable 3D representations \cite{božič2022neural, wu20234d, hasselgren2022nvdiffrecmc}. These methods employ signed distance functions (SDFs), meshes, or volumetric representations to model geometry, while also estimating underlying properties like diffuse albedo and specular parameters and their results can be used for relighting in novel environments. However, as these approaches are supervised on captured image data under a fixed illumination, there still exists an ambiguity between the underlying properties that images alone cannot explain. Multiple possible materials and textures could be composed onto the geometry to produce the same final images in the training data, leading to fundamental ambiguities when inferring underlying albedo and material properties using a simple pixel-level reconstruction loss. The ill-posed nature of this problem ultimately leads to suboptimal factorization. 

Our key insight is that 2D priors over plausible materials and albedos, in addition to reconstruction losses, can resolve ambiguities in factorized inverse rendering. 
We learn a large scale conditional diffusion prior over likely materials for RGB images under different illuminations. In addition to reconstruction loss, 
we distill the fine-tuned diffusion model to provide additional signal about plausible texture and material combinations for the depicted object during 3D optimization. 

We demonstrate our 3D inverse rendering approach, MaterialFusion, on the NeRF Synthetic, NeRFactor datasets~\cite{mildenhall2020nerf, zhang2021nerfactor}, a test set of our BlenderVault dataset and the Stanford-ORB dataset~\cite{kuang2023stanford}. We use these datasets to show significant improvements in novel view synthesis under relighting as well as material estimation compared to prior state-of-the-art work on both synthetic and real data. We trained a conditional diffusion model, StableMaterial, with albedos, materials, and relit images rendered from \raisebox{0.5ex}{\texttildelow}29K high quality objects, augmented with our own BlenderVault dataset of \raisebox{0.5ex}{\texttildelow}12K high quality synthetic Blender objects curated from online sources, and show its superior performance compared to previous approaches that recover albedo and materials from input images. Using our prior, our learned factorized representation generalizes better to novel lighting conditions across diverse lighting, object, and underlying material scenarios, as shown in the relighting results in Fig. \ref{fig:teaser}.
\section{Related Works}
\label{sec:related}

\subsection{Inverse Rendering}
In recent years, reconstruction methods that learn a 3D representation from a set of multi-view images have rapidly improved~\cite{mildenhall2020nerf, barron2023zipnerf, mueller2022instant, barron2022mipnerf360, kerbl20233d} in terms of quality and speed. However, many of these methods do not disentangle the underlying texture and materials, from the illumination. Therefore, rendering the acquired scene under novel lighting conditions remains a challenge.

To address this, inverse rendering works have begun focusing on reconstructing the 3D appearance along with the underlying materials of a scene or object. Given a set of images of a scene or object under a fixed illumination, some works have aimed to recover the texture, materials, and lighting~\cite{zhang2021nerfactor, hasselgren2022nvdiffrecmc, Jin2023TensoIR, liang2023gsir, zhang2022invrender, boss2021neuralpil, R3DG2023, Munkberg_2022_CVPR}. This task is inherently challenging due to its high dimensionality and ambiguity in explaining the image appearance, as multiple illumination and material parameter combinations can be used to reproduce the final appearance. To tackle this ambiguity, other works simplify the problem setting by assuming or modeling scene lighting~\cite{Jin2023TensoIR, iron-2022, Guo_2022_CVPR, cheng2023wildlight} or employing domain-specific priors~\cite{zhang2021nerfactor, Bi_2020_deep3d, chen2024urhand} to inject additional information on physical properties. Nevertheless, assumptions about lighting limit the applicability of these methods in real-world scenarios such as online marketplaces, where lighting conditions can be difficult to capture and are constantly changing. Moreover, the priors used in the aforementioned works are either trained on small-scale or procedurally generated data or focus on a specific object category.

In contrast, our approach does not rely on controlled lighting conditions; instead, it primarily utilizes a large-scale 2D texture and material prior trained with a large synthetic object dataset we curated.
The objects in this dataset contain complex Physically Based Rendering (PBR) assets, enhancing our prior's predictions.

\subsection{2D Diffusion Priors For 3D Tasks}
The success of diffusion models in text-to-image synthesis~\cite{rombach2021highresolution, saharia2022photorealistic, ho2020denoising} has also brought attention to employing large scale 2D priors for 3D generation \cite{poole2022dreamfusion,wang2023score,lin2023magic3d, Chen_2023fantasia3D,metzer2022latent,wang2023prolificdreamer, sun2023dreamcraft3d}. Dreamfusion \cite{poole2022dreamfusion} and SJC \cite{wang2023score} first propose Score Distillation Sampling (SDS) to optimize a 3D representation using 2D diffusion model gradients. 
Some follow-up works enriched the 2D model prior with 3D knowledge by fine-tuning the model to generate novel views of an object~\cite{liu2023zero1to3}, to generate images of several views simultaneously~\cite{wang2023imagedream, shi2023mvdream, liu2023syncdreamer}. Moreover, it has been shown that such enriched models perform better in generating 3D models from scratch and in single-view reconstruction~\cite{liu2023zero1to3, wang2023imagedream, shi2023mvdream, liu2023one2345, zhou2023sparsefusion, liu2023one2345++, long2023wonder3d, shi2023zero123, liu2023syncdreamer}.
Additionally, ReconFusion \cite{wu2023reconfusion} also uses 2D diffusion priors to improve sparse-view 3D reconstruction.
However, common to all of these works is the lack of material and illumination disentanglement, thereby limiting the relighting performance of the generated or reconstructed objects.

To predict physical properties, previous works showed success in finetuning a pretrained diffusion model. Specifically, some works predict material parameters given an RGB image~\cite{vecchio2023controlmat, MatFusion, zeng2024rgbtox, kocsis2023intrinsic}. However, these works reconstruct only a 2D representation of the underlying physical properties, and do not consider the 3D reconstruction from a set of images. In contrast to the aforementioned works, our approach reconstructs the underlying 3D geometry, material properties, and environmental lighting from a set of multi-view images via score distillation.
Closest to ours, ~\cite{chen2024intrinsicanything} concurrently used a 2D diffusion model to guide relightable 3D inference, but used diffusion samples to guide the optimization while ours uses likelihood maximization via SDS.

\subsection{3D Datasets with PBR assets}

The availability of 3D datasets is considerably smaller than the availability of 2D datasets, even more so in terms of PBR information, imposing a challenge in 3D-related tasks. In particular, commonly used 3D datasets~\cite{downs2022google, co3d, jampani2023navi} lack PBR information. 
Some datasets~\cite{photoshape, collins2022abo} offer 3D objects with PBR information but are limited in diversity to only furniture. Objaverse~\cite{deitke2023objaverse} offers diversity yet contains many objects that are partial reconstructions, low in quality, or cartoonish. Artist-designed high-quality 3D objects with PBR data are available in different sources, but are not organized in a dataset suitable for research. 
In this work, we introduce a new dataset of Blender objects containing high quality PBR assets curated from online sources. We use this dataset to augment previous datasets, greatly enhancing the diversity of PBR information available for training.
\begin{figure*}
    \centering
    \includegraphics[width=\textwidth]{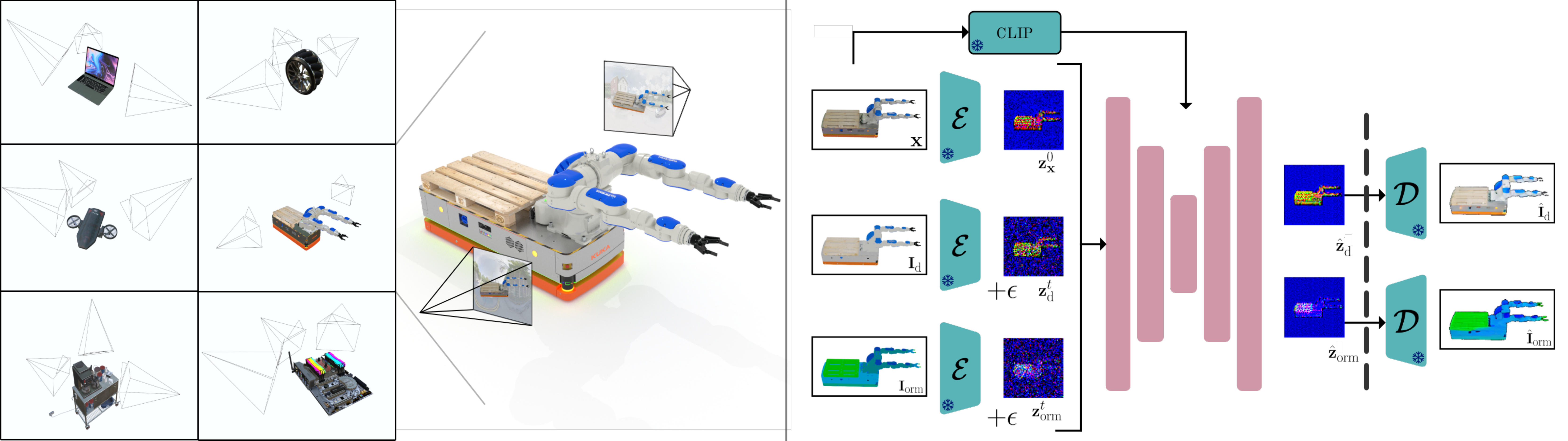}
    \caption{StableMaterial receives an RGB image as input and outputs the albedo $\hat{\mathbf{I}}_\text{d}$ and ORM $\hat{\mathbf{I}}_\text{orm}$ 2D maps. To train StableMaterial, we use BlenderVault objects to render a dataset of multi-view images under varying illuminations as well as the corresponding albedo and ORM maps. Given a triplet $(\mathbf{x}, \mathbf{I}_\text{d}, \mathbf{I}_\text{orm})$ of an image and its albedo and ORM maps, we encode them using the pretrained Stable Diffusion encoder and concatenate the image latent with the noisy albedo and ORM latents. The model is then trained with a diffusion loss to denoise the albedo and ORM maps.}
    \label{fig:2d_approach}
    \vspace*{-2mm}
\end{figure*}
\section{Methodology}
\label{sec:methodology}

\begin{figure*}
    \centering
    \hspace*{-1.4cm}
    \includegraphics[width=0.8\textwidth]{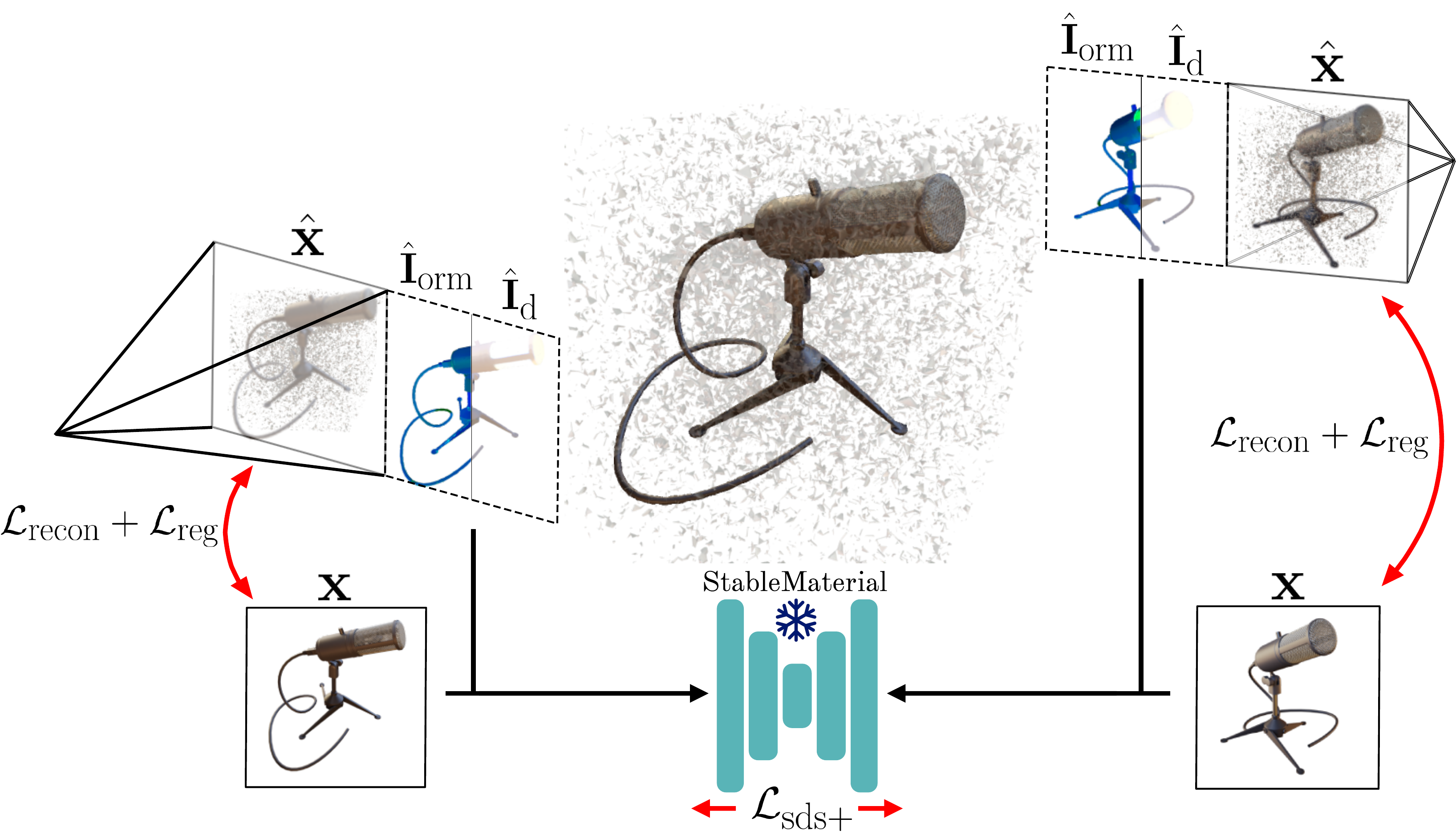}
    \caption{
    MaterialFusion reconstructs an object's geometry, PBR materials, and environmental illumination from a set of multi-view images under a fixed lighting condition. In addition to the reconstruction and regularization losses computed between our rendered images $\hat{\mathbf{x}}$ and reference RGB images $\mathbf{x}$, MaterialFusion employs priors from our pre-trained StableMaterial to enhance PBR material reconstruction. Specifically, it calculates an SDS loss for the rendered albedo and ORM components, $\hat{\mathbf{I}}_\text{d}$ and $\hat{\mathbf{I}}_{\text{orm}}$ conditioned on $\mathbf{x}$.
    }
    \label{fig:3d_approach}
    \vspace*{-4mm}
\end{figure*}

This section introduces MaterialFusion, our approach to reconstructing a 3D representation of an object from a set of multi-view images. An overview of our approach is shown in Fig.~\ref{fig:3d_approach}.
Specifically, given a set of posed images of the object captured under an unknown illumination, our goal is to reconstruct the object's geometry and BRDF appearance, as well as recover the environmental illumination. Accurately reconstructing these components allows us to faithfully recreate the object appearance under new lighting conditions.
We represent the geometry as a mesh, as its explicit nature is more suitable for downstream tasks.
For the material, we use
a simplified Disney principled BRDF model \cite{burley2012physically} representation. Specifically, the material texture contains three components per texel, albedo $\mathbf{a} \in \mathbb{R}^3$, roughness $r \in \mathbb{R}$, and metallicness $m \in \mathbb{R}$. Following prior works \cite{hasselgren2022nvdiffrecmc, Munkberg_2022_CVPR}, we represent the albedo texture as an albedo UV-map $\mathbf{k}_\text{d}$, and roughness and metallicness as part of an occlusion, roughness, metallicness (ORM) UV-map $\mathbf{k}_\text{orm}$, where each texel is $(o,r,m)$ with $o$ unused. The environment illumination is represented as a high dynamic range (HDR) environment map.

Our key idea is to leverage a strong 2D prior obtained from an image diffusion model which is trained to estimate the underlying material given a RGB image input.
To accomplish this, we first adapt an existing image diffusion model (Stable Diffusion 2.1~\cite{rombach2021highresolution}) to predict the albedo and ORM from an image of an arbitrary object rendered under a randomly selected illumination. This allows us to extend an existing 2D diffusion prior such that it has material understanding. The finetuning procedure is shown in Fig.~\ref{fig:2d_approach}.
We then leverage the extended 2D prior in an inverse rendering framework to infer a disentangled 3D representation of a given object and an HDR map of the environment lighting. Specifically, we utilize a variant of SDS loss~\cite{poole2022dreamfusion} to employ the 2D prior for 3D optimization. 
We show an overview of the 3D inference procedure in Fig.~\ref{fig:3d_approach}.

\subsection{Training Data}
Learning a diffusion prior for albedo and ORM prediction from images we leverage a diverse dataset of synthetic object renderings with high-quality PBR textures.
Using such data, we generate training images with a graphics engine capable of reproducing realistic appearances such as Blender. We examined existing datasets such as Objaverse~\cite{deitke2023objaverse} and ABO~\cite{collins2022abo} for this purpose.

Objaverse is a large and diverse dataset, but it contains many unrealistic, low-quality, or textureless objects. To address this, we followed a similar filtering procedure as~\cite{tang2024lgm}, and then further filtered for non-cartoon objects with PBR textures. This resulted in a filtered subset of \raisebox{0.5ex}{\texttildelow}8.5K objects from Objaverse.

While the filtered Objaverse dataset provided good coverage, we found that augmenting it with the ABO dataset (which contains \raisebox{0.5ex}{\texttildelow}8K objects from only 63 categories) was not sufficient to achieve the desired diversity in our training data. To further improve the diversity, we created our own BlenderVault dataset, which contains an additional \raisebox{0.5ex}{\texttildelow}12K high-quality, PBR-textured objects. BlenderVault consists of Blender objects designed and validated by artists across arbitrary categories for use in commercial projects.

To render the training images, we replaced any glass surfaces in the objects with a black surface of roughness 0.25 and metallicness 0. We then rendered 30 images of each object, with randomly selected azimuth~$\sim\lbrack0^\circ, 360^\circ\rbrack$ and elevation~$\sim\lbrack-15^\circ, 90^\circ\rbrack$ on a hemisphere with a radius~$\sim\lbrack1.5, 2.0\rbrack$. The lighting conditions were also varied, using a random selection of StreetLearn~\cite{mirowski2019streetlearn} environment maps, Laval~\cite{laval_hdr} indoor environment maps, point lights, or directional sun lights.

In total, our training dataset consists of \raisebox{0.5ex}{\texttildelow}28K synthetic objects with high quality PBR assets, combining the filtered Objaverse, ABO, and our own BlenderVault data.

\subsection{StableMaterial -- 2D Material Denoising Diffusion Prior}
To have a strong 2D material prior, we build on Stable Diffusion 2.1~\cite{rombach2021highresolution} and fine-tune it from the pretrained model on a dataset consisting of triplets $(\mathbf{x}, \mathbf{I}_\text{d}, \mathbf{I}_\text{orm})$, where $\mathbf{x}$ is an RGB image of the object, and $\mathbf{I}_\text{d}, \mathbf{I}_\text{orm}$ are its corresponding rendered albedo and ORM components, respectively. Formally, given an RGB input $\mathbf{x}$ of an object under unknown illumination, we fine-tune Stable Diffusion to output its underlying albedo $\mathbf{I}_\text{d}$ and ORM $\mathbf{I}_\text{orm}$ components.

\paragraph{Model Architecture.}
We modify only Stable Diffusion's UNet, so that it is conditioned on the input image $\mathbf{x}$ in two ways. First, we encode it with Stable Diffusion's pre-trained frozen VAE $\mathcal{E}$ and concatenate the resulting clean latent $\mathbf{z}^0_\mathbf{x}$ to the noisy latent codes $\mathbf{z}^t$ in the channel dimension, where $t$ is the diffusion timestep. Specifically, the noisy latent code $\mathbf{z}^t = [\mathbf{z}_\text{d}^t, \mathbf{z}_\text{orm}^t]$, i.e. the concatenation of the noisy albedo latent $\mathbf{z}_\text{d}^t$ and the noisy ORM map $\mathbf{z}_\text{orm}^t$ in the channel dimension. The input of our UNet is then $\left(\mathbf{z}^0_\mathbf{x}, \mathbf{z}^t, t \right)$. The text conditioning is also replaced with a CLIP image embedding of the input image. These two ways of inputting the image into the model allow it to have both global and local reasoning about the image.

To output both albedo and ORM maps, our noisy latent codes $\mathbf{z}^t$ consist of 8 channels, 4 corresponding to the albedo and the other 4 corresponding to the ORM.
In total, the input of our network is composed of 12 channels consisting of the encoded input image, noisy albedo latents and the noisy ORM latents. To obtain the RGB albedo and ORM maps we decode the denoised $\hat{\mathbf{z}}$ through the pre-trained Stable Diffusion decoder $\mathcal{D}$. To account for the different input and output channels, the first and last layers are changed and randomly initialized, while the other UNet parameters are kept unchanged. 

\paragraph{Loss.}

To fine-tune the model, we utilize v-prediction diffusion loss~\cite{salimans2021progressive}.
At each training iteration, we sample a triplet $(\mathbf{x}, \mathbf{I}_\text{d}, \mathbf{I}_\text{orm})$, and encode each of the images with $\mathcal{E}$. We concatenate $\mathcal{E}(\mathbf{I}_\text{d}), \mathcal{E}(\mathbf{I}_\text{orm})$ in the channel dimension and denote their concatenation by $\mathbf{z}$.
We sample a diffusion timestep $t$ along with an 8-channel random noise $\epsilon$, and add the noise to $\mathbf{z}$ to obtain $\mathbf{z}^t$. 
The diffusion loss is defined as
\begin{equation}
    \mathcal{L}_{\text{diff}} = 
    \mathbb{E}_{\mathbf{x}, \mathbf{k}_\text{d}, \mathbf{k}_\text{orm}, \epsilon \sim \mathcal{N}(0, I), t}
    \left[
    \| \epsilon_\theta\left(\mathcal{E}(\mathbf{x}), \mathbf{z}^t, t \right) - \mathbf{v}_t \|_2^2
    \right],
\end{equation}
where $\mathbf{x}, \mathbf{z}^t$ are as defined above.
As in \cite{salimans2021progressive}, $\mathbf{v}_t = \alpha_t \epsilon - \sigma_t \mathbf{z}$, where $\alpha_t, \sigma_t$ are the parameters of the scheduler. Similarly to~\cite{brooks2022instructpix2pix}, we enable classifier-free guidance by setting the input images, input image prompt, or both to all zeros with a 5\% probability each and set the guidance scale to 3.0. 

\subsection{Prior-guided Inverse Rendering}

Having material knowledge in our trained StableMaterial model, we can distill this knowledge and reconstruct a 3D disentangled representation of the object. Specifically, given a set of multi-view images under an unknown illumination depicting an object, we aim to reconstruct the underlying geometry represented as a mesh and denoted by $\mathbf{G}$, the underlying UV material texture denoted by $(\mathbf{k}_\text{d}, \mathbf{k}_\text{orm})$, and the environment illumination $\mathbf{L}$. To this end, we directly optimize these representations and build on recent advancements in the distillation of 3D information from 2D diffusion models~\cite{poole2022dreamfusion}, in conjunction with the off the shelf nvdiffrecmc inverse rendering pipeline~\cite{hasselgren2022nvdiffrecmc} as part of MaterialFusion.

Following previous works~\cite{hasselgren2022nvdiffrecmc, Munkberg_2022_CVPR} we parameterize the geometry through an SDF denoted by $\mathbf{S}$, and extract a mesh $\mathbf{G}$ in each optimization iteration using DMTet~\cite{NEURIPS2021_30a237d1}. 
Given the mesh $\mathbf{G}$, the texture $(\mathbf{k}_\text{d}, \mathbf{k}_\text{orm})$, a camera view $\mathbf{C}$, and an HDR environment light map $\mathbf{L}$, we use nvdiffrecmc's differentiable renderer to produce a 2D rendering. 

In each optimization iteration, we sample some images and their associated views $\mathbf{x}, \mathcal{C}$, respectively, from the training set, and differentiably render the object using the optimized parameters from the views $\mathcal{C}$. We obtain the rendered image $\hat{\mathbf{x}}$ with nvdiffrecmc's renderer and apply a reconstruction loss to optimize $(\mathbf{S}, \mathbf{k}_\text{d}, \mathbf{k}_\text{orm}, \mathbf{L})$:
\begin{equation}
    \mathcal{L}_\text{recon} = \mathbb{E}_{\mathbf{C}} \left[ \mathcal{L}_2( \hat{\mathbf{x}},  \mathbf{x} ) \right],
\end{equation}
\begin{figure*}
    \centering
    \includegraphics[trim={0 9.5cm 0 9.5cm},clip,width=\textwidth]{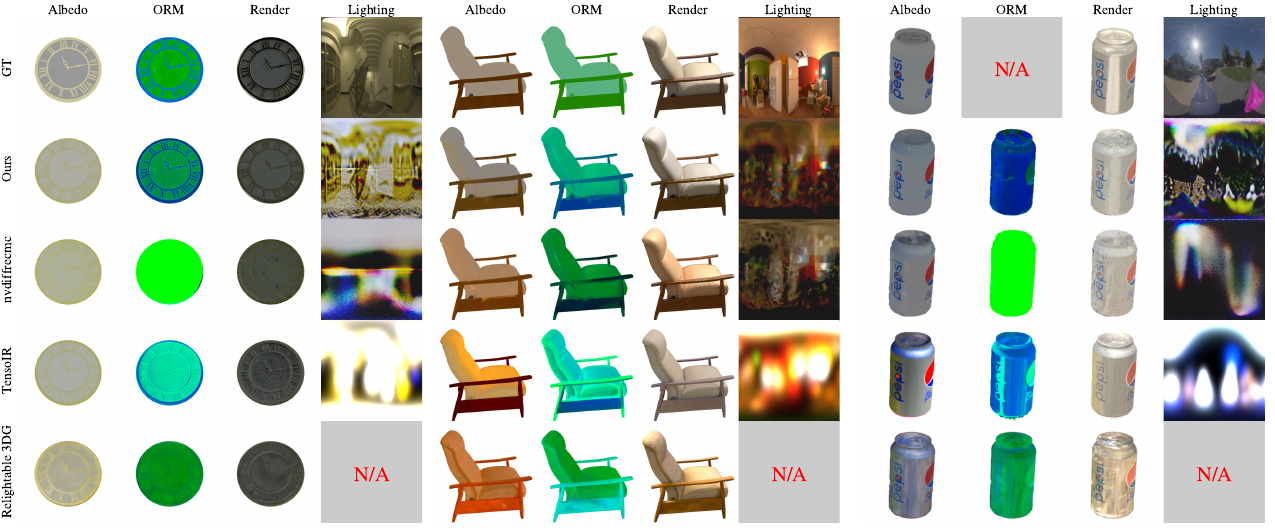}
    \caption{Qualitative comparison for MaterialFusion vs. other methods. We present the 3D reconstructed albedo, ORM, environment light map, and relit rendered images for three different objects, both synthetic and real. Our method demonstrates better accuracy compared to the baseline methods, as can be seen by the accuracy of the reconstructed materials and the relit image appearance. Our prior also acts as an additional regularizer on other 3D properties such as geometry and illumination.}
    \label{fig:3d_comparison}
\vspace*{-3mm}
\end{figure*}

At the same time, we also render the corresponding components of the albedo and the material, $\hat{\mathbf{I}}_\text{d}$ and $\hat{\mathbf{I}}_{\text{orm}}$, respectively, encode them with the Stable Diffusion model encoder and concatenate their latent representations to get $\mathbf{z}$. 
Then, we sample a Gaussian noise $\epsilon$ and add it to $\mathbf{z}$ according to a random diffusion timestep $t\in[0.02, 0.98]$, to obtain $\mathbf{z}^t$. We then denoise $\mathbf{z}^t$ using the diffusion model, and sample a clean latent representation of the materials denoted by $\hat{\mathbf{z}}$ via DDIM sampling for 5 steps, conditioned on $\mathbf{x}$. 
We use the SDS loss in both latent and pixel space:
\begin{equation}
    \mathcal{L}_\text{SDS+} = \mathbb{E}_{t, \epsilon, v} \left[\lambda_{\text{latent}}||\mathbf{z} - \hat{\mathbf{z}}||^2 + \lambda_{\text{rgb}}||\mathcal{D}(\mathbf{z}) - \mathcal{D}(\hat{\mathbf{z}})||^2
    \right],
\end{equation}
where $\mathcal{D}$ is the Stable Diffusion decoder. This loss is inspired by HiFA~\cite{zhu2023hifa}, as we empirically found the RGB term important for boosting the quality of materials estimated during training, shown in Tab.~\ref{tab:ablations}. 
Using 5 denoising steps was key for producing crisp and accurate predictions for the albedo and ORM materials. Finally, our total loss consists of the reconstruction loss, regularization loss, and SDS loss:

\begin{equation}
    \mathcal{L}_\text{MaterialFusion} = \mathcal{L}_{\text{recon}} + \mathcal{L}_{\text{reg}} + \gamma_i\mathcal{L}_{\text{SDS+}},
\end{equation}
where $\gamma_i$ is an iteration dependent hyperparameter which decays as training progresses. We find that reducing the weight on SDS loss towards the end of the optimization helps preserve finer details that may be lost due to the encoder/decoder operation. We use the same $\mathcal{L}_{\text{reg}}$ as nvdiffrecmc~\cite{hasselgren2022nvdiffrecmc}. Conceptually, using SDS loss during inverse rendering maximizes the likelihood of the ORM and albedo under our prior for all training images. 
\section{Experiments}
\label{sec:experiments}

We evaluate MaterialFusion and StableMaterial on image sequences of objects made of various materials and textures and show the qualitative and quantitative comparisons. We first evaluate MaterialFusion against prior inverse rendering methods for object relighting on a number of synthetic and real diverse objects and highlight the advantages of our approach in terms of appearance relighting. We further compare the trained 2D prior against other previous approaches that predict albedo and material from a single image using test data from BlenderVault excluded from training. 

\subsection{Relightable 3D Reconstruction}
For MaterialFusion, we adopt a validation setup similar to nvdiffrecmc by relighting the objects under novel illuminations and then comparing to the groundtruth relit object images. For synthetic objects, we acquire the groundtruth relightings by rendering images of synthetic objects under unseen illuminations, while real objects are captured in novel environments for which the illuminations are computed. 
\begin{table*}
    \centering
    \resizebox{\linewidth}{!}{\begin{tabular}{lccc  ccc  ccc ccc}\toprule
    & \multicolumn{3}{c}{NeRF Synthetic} & \multicolumn{3}{c}{NeRFactor} & \multicolumn{3}{c}{BlenderVault} & \multicolumn{3}{c}{Stanford-ORB} \\\cmidrule(lr){2-4}\cmidrule(lr){5-7}\cmidrule(lr){8-10}\cmidrule(lr){11-13}
    & PSNR $\uparrow$ & SSIM $\uparrow$ & LPIPS $\downarrow$ & PSNR $\uparrow$ & SSIM $\uparrow$ & LPIPS $\downarrow$ & PSNR $\uparrow$ & SSIM $\uparrow$ & LPIPS $\downarrow$ & PSNR $\uparrow$ & SSIM $\uparrow$ & LPIPS $\downarrow$\\ \hline
nvdiffrecmc    & \cellcolor{tabsecond}25.70 & \cellcolor{tabsecond}0.924 & \cellcolor{tabsecond}0.090 & \cellcolor{tabsecond}25.91 & \cellcolor{tabsecond}0.921 &  \cellcolor{tabthird}0.092 & \cellcolor{tabsecond}25.11 & \cellcolor{tabsecond}0.910 &                      0.167 &
\cellcolor{tabsecond}31.10 & \cellcolor{tabfirst}0.968 & \cellcolor{tabthird}0.048\\
TensoIR        &  \cellcolor{tabthird}24.32 &  \cellcolor{tabthird}0.923 & \cellcolor{tabsecond}0.090 &  \cellcolor{tabthird}24.87 &  \cellcolor{tabthird}0.916 &                      0.094 &  \cellcolor{tabthird}25.01 &                      0.903 &  \cellcolor{tabthird}0.162 &
\cellcolor{tabthird}28.81 & \cellcolor{tabthird}0.959 & \cellcolor{tabsecond}0.047\\
Relightable3DG &                      23.08 &                      0.897 &  \cellcolor{tabthird}0.094 &                      23.99 &                      0.908 &  \cellcolor{tabfirst}0.082 &                      22.83 &  \cellcolor{tabthird}0.909 & \cellcolor{tabsecond}0.148 &
27.40 & 0.955 & \cellcolor{tabthird}0.048\\

MaterialFusion           &  \cellcolor{tabfirst}26.26 &  \cellcolor{tabfirst}0.927 &  \cellcolor{tabfirst}0.085 &  \cellcolor{tabfirst}26.31 &  \cellcolor{tabfirst}0.922 & \cellcolor{tabsecond}0.091 &  \cellcolor{tabfirst}26.33 &  \cellcolor{tabfirst}0.921 &  \cellcolor{tabfirst}0.143&
\cellcolor{tabfirst}31.68 & \cellcolor{tabsecond}0.967 & \cellcolor{tabfirst}0.046\\
        \bottomrule
    \end{tabular}
    }
    \caption{
    Comparison of novel view synthesis relighting. In each column, the \colorbox{tabfirst}{best}, \colorbox{tabsecond}{second best}, and \colorbox{tabthird}{third best} results are marked.
\vspace*{-4mm}
    }
    \label{tab:3d_results}
\end{table*}

\paragraph{Datasets.}

We use 4 objects from NeRFactor \cite{zhang2021nerfactor}, 5 objects from the NeRF synthetic dataset \cite{mildenhall2020nerf}, 9 test objects from BlenderVault, and 14 objects from the Stanford-ORB~\cite{kuang2023stanford} datasets. The first three datasets consist of diverse synthetic objects with camera poses and their groundtruth data allows for us to re-render and compare objects with different illuminations. The NeRFactor and BlenderVault objects are relit by eight low resolution environment maps while NeRF synthetic objects are relit by four high resolution environment maps, and the quality comparison is computed on a test set of eight unseen poses per environment map. We also show our relighting performance on real objects from the Stanford-ORB dataset, which has images, corresponding poses, and groundtruth illuminations allowing us to re-render and relight objects. Objects are relit under two novel illuminations and the relighting comparison is done using a test set of unseen poses per environment map.

\paragraph{Metrics.} The final results for the 3D pipeline relighting comparison are the average PSNR, SSIM, and LPIPS across all relighting test views for each dataset. The metrics for the albedo used were PSNR, SSIM, and L1, and PSNR and L1 for ORM. LPIPS was excluded for both since perceptual similarity does not matter for ORM, and the VGG network likely has not seen albedo images. Given the scaling ambiguity between the albedo and light intensity during inference, the channels of RGB and albedo images are scaled against groundtruth during validation~\cite{hasselgren2022nvdiffrecmc}. Since the ORM and albedo are fundamentally pixel-wise material parameters, we use the L1 metric to measure physical similarity.

\paragraph{Baselines.}
We compare MaterialFusion against three current state-of-the-art inverse rendering methods that estimate geometry, albedo, roughness, and metallicness from a set of images. These approaches are nvdiffrecmc \cite{hasselgren2022nvdiffrecmc}, which is the method our pipeline is built upon, Relightable 3D Gaussian \cite{R3DG2023}, and TensoIR \cite{Jin2023TensoIR}. 

\begin{table}
    \centering
    \resizebox{\linewidth}{!}{
    \begin{tabular}{lccc  cc}\toprule
    & \multicolumn{3}{c}{Albedo} & \multicolumn{2}{c}{ORM} \\
    \cmidrule(lr){2-4}\cmidrule(lr){5-6} & PSNR $\uparrow$ & SSIM $\uparrow$ & L1 $\downarrow$ & PSNR $\uparrow$& L1 $\downarrow$  \\ \hline
nvdiffrecmc    & \cellcolor{tabsecond}28.24 & \cellcolor{tabsecond}0.939 & \cellcolor{tabsecond}0.021 & \cellcolor{tabthird}15.93 & \cellcolor{tabthird}0.062\\
TensoIR        &  \cellcolor{tabthird}25.82 &  \cellcolor{tabthird}0.927 & \cellcolor{tabthird}0.026 & 14.19 & 0.063\\
Relightable3DG & 24.30 & 0.925 & 0.036 & \cellcolor{tabsecond}20.96 & \cellcolor{tabsecond}0.041\\
MaterialFusion           &  \cellcolor{tabfirst}29.31 & \cellcolor{tabfirst}0.949 & \cellcolor{tabfirst}0.015 & \cellcolor{tabfirst}22.21 & \cellcolor{tabfirst}0.033\\
        \bottomrule
    \end{tabular}
    }
    \caption{
    Reconstructed 3D albedo and ORM comparison on BlenderVault objects.
    }
\vspace*{-2mm}
    \label{tab:3d_albedo_orm_results}
\end{table}

\paragraph{Results.}

We present qualitative and quantitative results for both albedo and ORM estimation quality as well as performance during relighting. Fig.~\ref{fig:3d_comparison} shows a visual comparison of the albedo and ORM estimated by all methods. Our method is able to recover high frequency details in both the albedo and ORM that other methods are not able to. This in turn leads to better performance under novel relighting, where Tab.~\ref{tab:3d_results} shows our method achieving the highest scores across all three datasets. The source of improvement in the relighting performance is best understood via the results in Tab.~\ref{tab:3d_albedo_orm_results}, where the estimated albedo and ORM quality for BlenderVault objects were directly compared to the groundtruth. We were unable to compare for the NeRF and NeRFactor datasets as the underlying material shaders used did not conform directly to the albedo and ORM. 

The comparisons show that our method performs best in both albedo and ORM estimation, as other methods suffer from poor albedo or ORM estimates, leading to poorer relighting. The synthetic example of the clock in Fig.~\ref{fig:3d_comparison} shows how MaterialFusion is able to accurately disambiguate different areas of the material and albedo, leading to a much more accurate rendering under novel illumination where details aren't lost like in the other methods' renderings. This can also be seen in the real can example where our method accurately deduces it is metallic (shown by the strength of the blue channel in the ORM map) and is able to accurately replicate the reflection of the can similarly to the real world. Our method shows better semantic material understanding as it is able to correctly distinguish between different parts of an object that are made of different materials, leading to better decoupling between the reflectance and environment illumination. Our results confirm our prior's improvements against baselines by better inferring underlying physical properties on synthetic and real data.

\begin{figure*}
    \centering
    \includegraphics[trim={0 7.925cm 0 7.51cm},clip,width=0.93\textwidth]{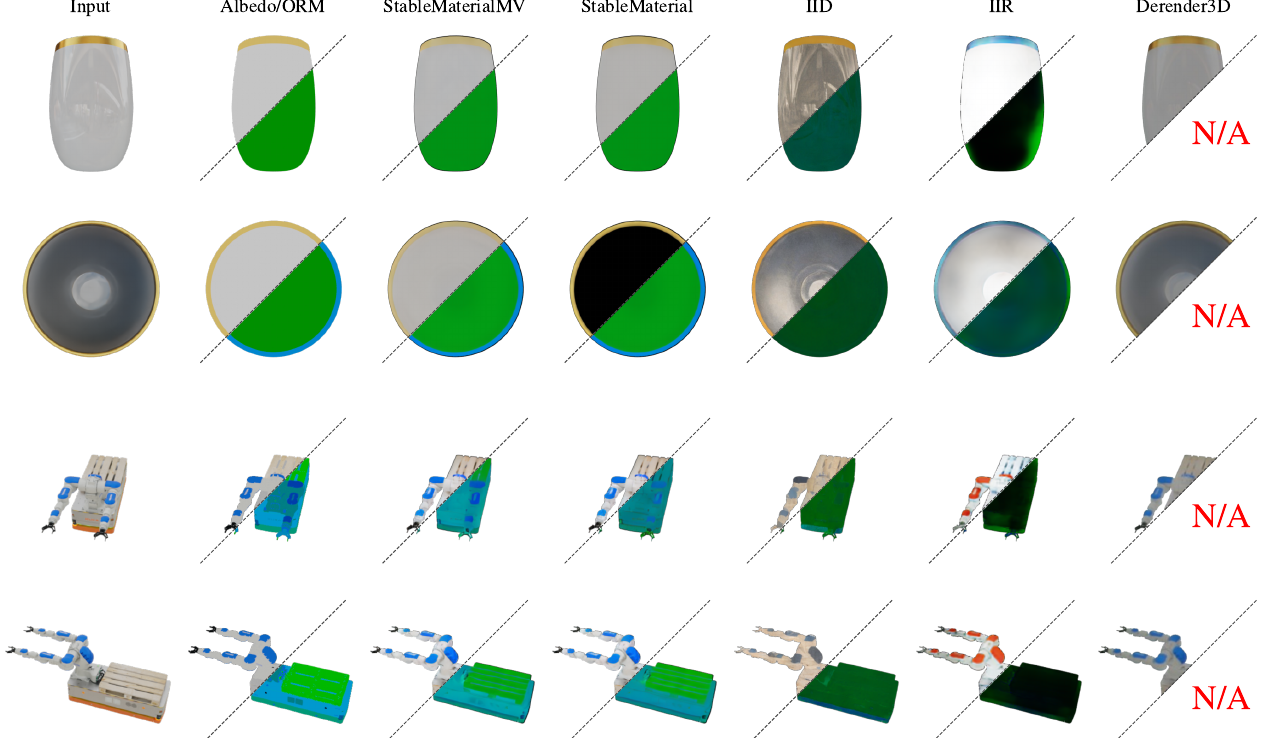}
\vspace*{-2mm}
    \caption{Qualitative comparison of the albedo and ORM 2D predictions. The Derender3D ORM data is marked as N/A since it does not offer ORM predictions. Given 4 images of an object, StableMaterial recovers complex material data. StableMaterialMV attends to appearance details across views, recovering consistent and high quality materials across challenging views, as seen in the cup example.
    }
\vspace*{-5mm}
    \label{fig:2d_comparison}
\end{figure*}

\subsection{Validating Material Inference from a 2D Input}
To validate StableMaterial's performance, we evaluate its performance on RGB images of synthetic test objects captured under an unknown illumination. We then directly compare the predictions to the groundtruth albedo and materials using extracted data from the synthetic objects. 
\begin{table}
    \centering 
    \resizebox{\linewidth}{!}{
    \begin{tabular}{lccc  cc}\toprule
    & \multicolumn{3}{c}{Albedo} & \multicolumn{2}{c}{ORM} \\
    \cmidrule(lr){2-4}\cmidrule(lr){5-6}
    & PSNR $\uparrow$ & SSIM $\uparrow$ & L1 $\downarrow$ & PSNR $\uparrow$ & L1 $\downarrow$ \\ \hline
Derender3D                 &                      21.69 &                      0.874 &                      \cellcolor{tabsecond}0.025 &                      -- &                      --  \\
IIR  &  \cellcolor{tabthird}22.62 & \cellcolor{tabsecond}0.905 &  \cellcolor{tabthird}0.026 &  18.68 &  \cellcolor{tabthird}0.041 \\
IID  &                      21.71 &                      0.871 &                      0.031 &                      \cellcolor{tabthird}18.87 &                       0.045 \\
StableMaterial                      & \cellcolor{tabsecond}24.25 &  \cellcolor{tabthird}0.902 &  \cellcolor{tabfirst}0.018 & \cellcolor{tabsecond}24.86 & \cellcolor{tabsecond}0.029 \\
StableMaterialMV        &  \cellcolor{tabfirst}24.70 &  \cellcolor{tabfirst}0.907 & \cellcolor{tabfirst}0.018 &  \cellcolor{tabfirst}26.34 &   \cellcolor{tabfirst}0.014\\
    \bottomrule 
    \end{tabular}
    }
    \caption{
            Comparison of albedo and ORM 2D predictions produced by our method versus other methods. We use the mean of 10 samples generated by StableMaterial and StableMaterialMV for the evaluation, where StableMaterialMV denotes our prior utilizing multi-view attention during inference. 
\vspace*{-2mm}
            }
    \label{tab:2d_results}
\end{table}

\paragraph{Datasets.} 
We utilize 8 diverse test objects from our BlenderVault datasets whose material data was not seen during training. We then render the groundtruth albedo, ORM, and RGB appearance under an unknown randomly selected fixed illumination for 4 views. This is done per each object.

\paragraph{Metrics.}
To account for the variance in StableMaterial's outputs, we follow the procedure described in \cite{kocsis2023intrinsic} and compute 10 estimates for the albedo and ORM images and average them together before comparing to the groundtruth. We further account for the scale ambiguity in the resulting albedo for all the baselines by rescaling to the groundtruth albedo. Similarly to the 3D evaluation, the metrics used for the albedo were PSNR, SSIM, and L1, and PSNR and L1 for ORM. The final results are computed as the mean across views for all objects.

\paragraph{Baselines.} We test StableMaterial against Inverse Indoor Rendering (IIR)~\cite{zhu2022learningbased} and Intrinsic Image Diffusion (IID)~\cite{kocsis2023intrinsic}, which were trained on scene data to directly predict the albedo, roughness, metallicness given a single image. We also include \cite{wimbauer2022derendering}, which was trained on diverse data and predicts the albedo but not materials. 

\paragraph{Multi-view Attention at Inference.}
To make StableMaterial produce material outputs consistent across 2D views, we follow previous works~\cite{wang2023imagedream, shi2023mvdream} and incorporate multi-view attention. Specifically, we input a batch of 4 images, and modify the self-attention layers of the model so that each latent pixel in each of the images attends to the latent pixels of all other images. As such, StableMaterial predicts the most likely material given all input image appearances. The self-attention layers of the network process the 4 images as a single large image, while the other layers process them independently. Importantly, this multi-view attention mechanism is only employed during inference for 2D images. We show that using multi-view attention improves the quality of inferred materials against other baselines.

\paragraph{Results.}
As shown in Tab.~\ref{tab:2d_results}, our trained model shows strong performance across multiple objects in estimating the Albedo and ORM quality. Notably, performance jumps further when multi-attention is used across 4 views, as our model can handle difficult views that offer little appearance information during inference. Fig.~\ref{fig:2d_comparison} shows a qualitative comparison of our model against previous approaches.

The consistent improvements in both 2D and 3D tasks highlight the effectiveness of our approach in capturing the underlying physical properties of objects. The multi-view variant further demonstrates the benefits of leveraging additional viewpoints to enhance the albedo and ORM prediction quality for difficult 2D views. We found no significant differences when employing multi-view attention during inverse rendering, given that the albedo and ORM representations are already optimized to be multi-view consistent. However, the performance boost for 2D images raises the potential for usage in sparse view scenarios or where 3D reconstruction is not needed or infeasible. 

\begin{table}
    \centering
    \resizebox{0.75\linewidth}{!}{\begin{tabular}{lccc}\toprule
    & \multicolumn{3}{c}{BlenderVault} \\\cmidrule(lr){2-4}
    & PSNR $\uparrow$ & SSIM $\uparrow$ & LPIPS $\downarrow$ \\ \hline
$\gamma_i = 1$           &  25.50 &  0.916 &  0.150\\
$\lambda_\text{RGB} = 0$           &  21.41 &  0.871 &  0.233\\
$\lambda_\text{latent} = 0$           &  26.12 &  0.917 &  0.147\\
Ours & 26.33 & 0.921 & 0.143\\
        \bottomrule
    \end{tabular}}
    \caption{
    Effects of ablating elements from $\mathcal{L}_\text{MaterialFusion}$.}
    \label{tab:ablations}
\vspace*{-6mm}
\end{table}
\subsection{Ablation Studies}

In Tab.~\ref{tab:ablations}, we ablate three terms of ~$\mathcal{L}_\text{MaterialFusion}$ and evaluate relighting performance on the BlenderVault test dataset. All other parameters are unchanged when ablating one parameter. Setting $\lambda_\text{RGB} = 0$ particularly affects performance; by backpropagating through the SD encoder, the latent SDS term gradient introduces artifacts in the materials, degrading their quality. We also conduct an ablation where $\gamma_i$ is set to 1 throughout the inverse rendering optimization. This leads to a noticeable drop in performance, as materials estimated for objects with finer details suffer. 
\vspace*{-2mm}
\section{Conclusion}
\label{sec:conclusions}

In this paper, we introduced MaterialFusion, a 3D inverse rendering approach that utilizes StableMaterial, a 2D diffusion model finetuned from Stable Diffusion as a prior for enhancing the underlying materials during training. We introduced BlenderVault, a dataset of high quality objects and underlying PBR assets used to finetune our prior, enabling it with knowledge to recreate complex materials from images. Utilizing our prior on top of an off the shelf inverse rendering approach lead to a significant performance boost when for inferring relightable 3D representations. While our work introduces distillation of material knowledge in a 3D scenario, we believe there is great potential in utilizing our prior for applications in 2D or sparse-view settings. 

\section{Acknowledgments}
\label{sec:acknowledgments}
The authors would like to thank Jianjin Xu and the rest of the students from the Human Sensing Laboratory for their helpful feedback. This work was supported in part by the NSF GFRP (Grant No. DGE1745016) and NSF Award IIS-2345610.

{
    \small
    \bibliographystyle{ieeenat_fullname}
    \bibliography{main}
}
\clearpage
\setcounter{page}{1}
\maketitlesupplementary

\label{sec:supplementary}

\section{Additional Visualizations}
We show visualizations of StableMaterial and MaterialFusion's performance on additional examples from the BlenderVault test dataset as well as the NeRFactor and NeRF synthetic datasets. A webpage containing videos of the material reconstruction and relighting results has also been attached.
\section{Additional Details}
\paragraph{BlenderVault.}
To collect BlenderVault, we utilized BlenderProc~\cite{Denninger2023}to download objects from the BlenderKit website.  We use Blender in order to render out 30 $512\times512$ multi-view images, specifically using Cycles engine with 64 SPP. At each rendering operation, we load in the lighting by randomly selecting between three options with equal probability: StreetLearn environment maps that were reconstructed into HDR maps~\cite{Santos_2020}, Laval indoor environment maps, or a Blender light source. If a Blender light source is picked, then either a point light is set up at the camera location with 150W power, or a sun light points codirectionally with the camera at the object, with power amount randomly sampled between 10 and 20. To accurately render the albedo and ORM parameters of objects, we tried to approximate them as a Disney principled BRDF model as best as possible, corresponding to the ''Base Color'' and roughness and metalness parameters of the principled BSDF. Due to the diversity of shaders used to represent materials, dome objects had material or albedo that couldn't be rendered due to either the complexity of the object, size ($>$500MB), or due to interference from features such as procedural generation. In such cases we skip the objects and continue to rendering the next one. Overall, around 300 objects were skipped but are still included in the final dataset. 

\paragraph{Training Details.}

StableMaterial was trained in similar fashion to Zero123, by using a batch size of 1536 with images resized to 256$\times$256 and learning rate of $10^{-4}$ with an AdamW optimizer for 25k steps. A fully connected layer $(1028\rightarrow 1024)$ that converts the concatenation of the CLIP embedding and pose to a compatible embedding for Stable Diffusion's UNet was trained with a learning rate of $10^{-3}$, where the pose representation used was similar Zero123's. Stable Diffusion's UNet and the fully connected layers were trained and all other components were frozen. Finetuning took 2 days with 8$\times$ H100 GPUs. The filtering keys used for including Objaverse data were [\texttt{pbr}, \texttt{pbrtexture}, \texttt{substance}, \texttt{substancepainter}], while excluding objects that included [\texttt{style}, \texttt{stylized}, \texttt{cartoon}, \texttt{lowpoly}, \texttt{poly}]. These objects were filtered from the dataset gathered by~\cite{tang2024lgm}. 

\begin{figure*}
    \centering
    \includegraphics[trim={0 3.5cm 0 3.5cm},clip,width=\textwidth]{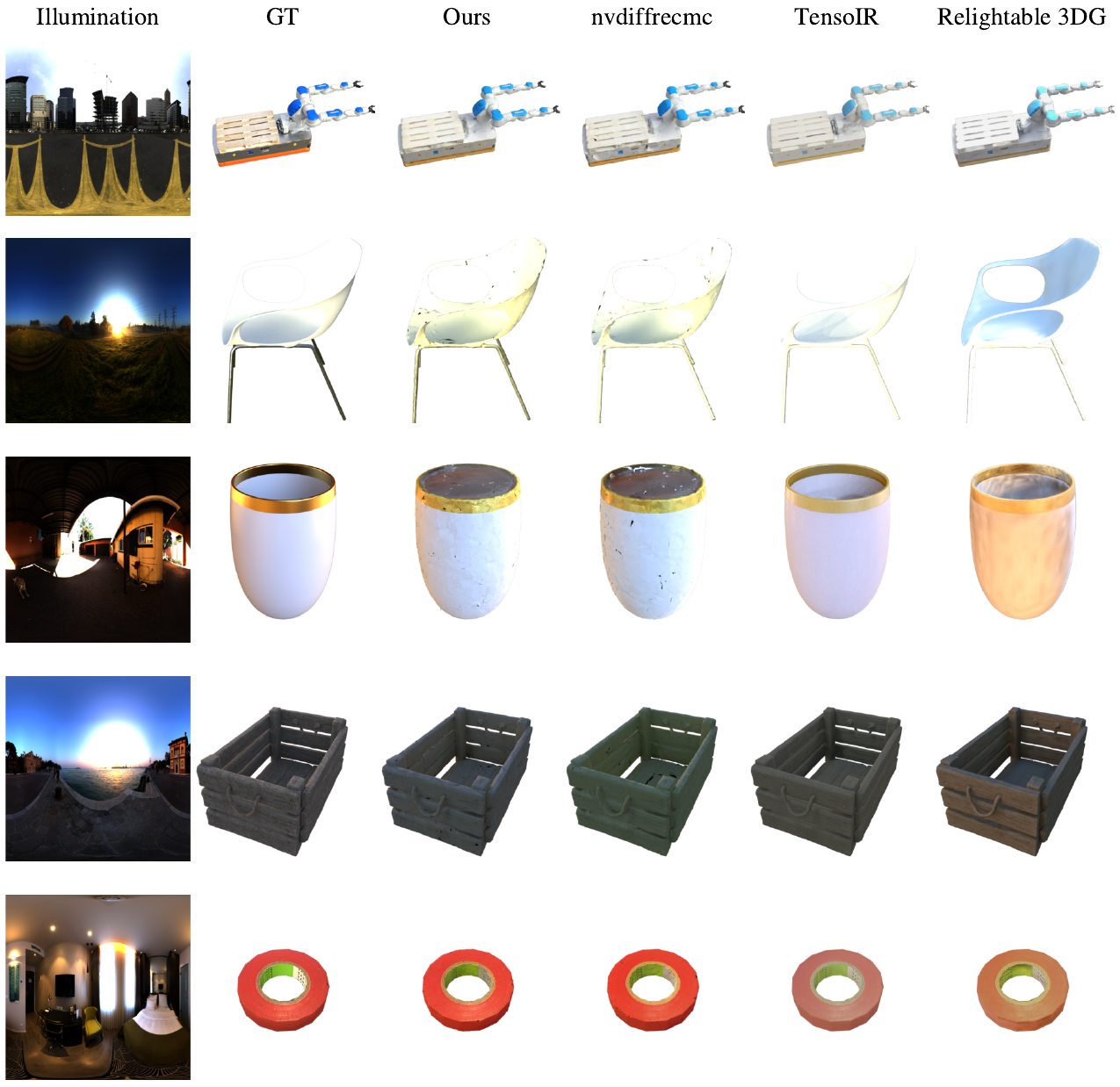}
    \caption{Additional comparisons for MaterialFusion vs. other methods on 3D physical properties reconstruction on more objects from the BlenderVault test dataset.}
    \label{fig:3d_comparison_3}
\end{figure*}

\begin{figure*}
    \centering
    \includegraphics[trim={0 6cm 0 5.5cm},clip,width=\textwidth]{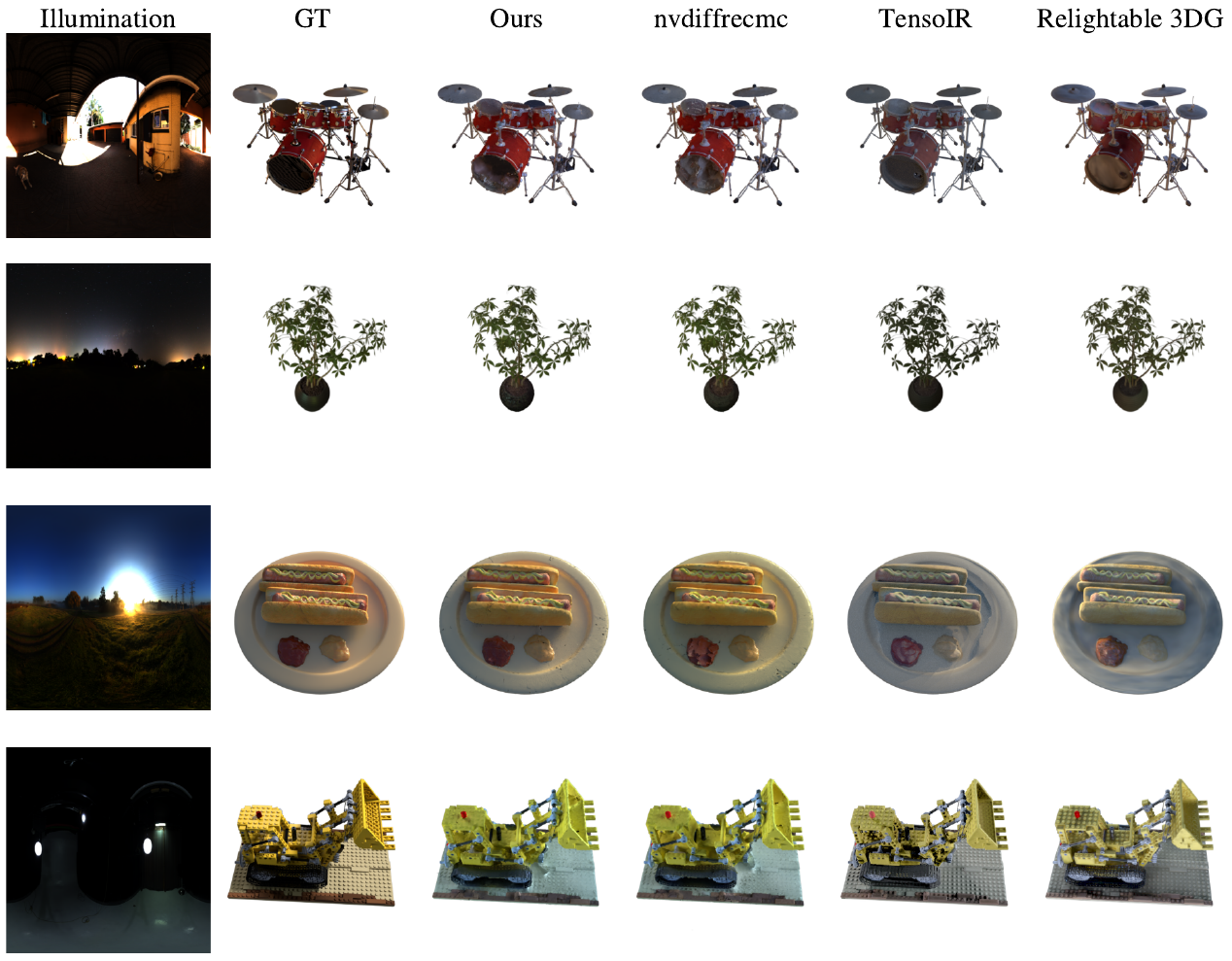}
    \caption{Additional comparisons for MaterialFusion vs. other methods on 3D physical properties reconstruction on the NeRFactor dataset.}
    \label{fig:3d_comparison_4}
\end{figure*}

\begin{figure*}
    \centering
    \includegraphics[trim={0 3.5cm 0 3.5cm},clip,width=\textwidth]{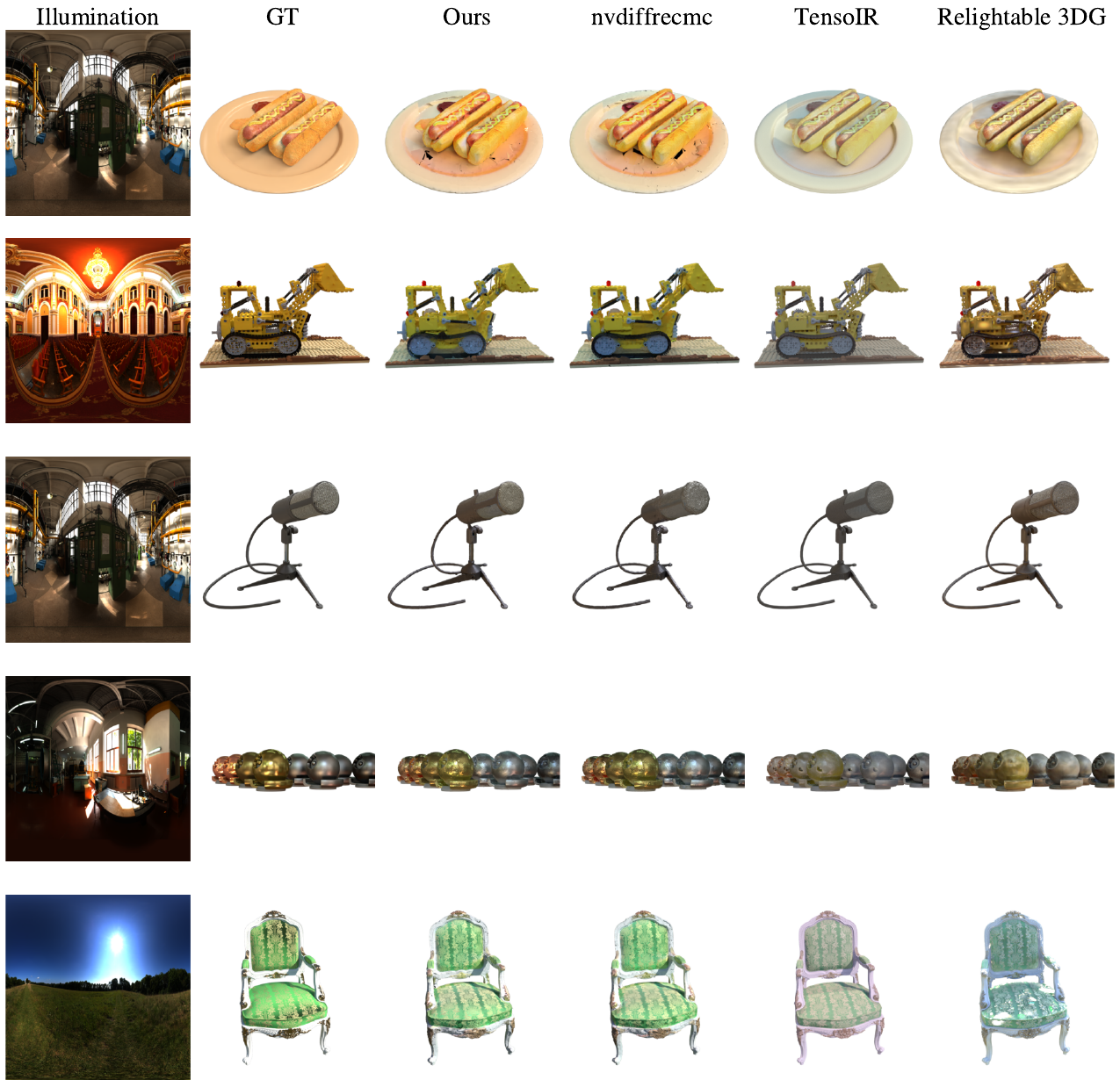}
    \caption{Additional comparisons for MaterialFusion vs. other methods on 3D physical properties reconstruction on the NeRF dataset.}
    \label{fig:3d_comparison_5}
\end{figure*}

MaterialFusion utilizes the nvdiffrast~\cite{Laine2020diffrast} differentiable renderer in order to render out the appearance. nvdiffrast can also render other properties for the system, such as the albedo, ORM, diffuse and specular lighting, and more. In particular, we feed the rasterized estimated albedo and ORM batches to StableMaterial. $\mathcal{L}_\text{recon}$ required sRGB tonemapped image inputs, and we took care to make sure that the inputs to StableMaterial were converted to RGB. $\mathcal{L}_\text{reg}$ contains a normal, albedo, and ORM smoothness regularizers, as well as a normal perturbation regularizer that encourages normal map perturbations. The last term included is a demodulated lighting regularization term that utilizes the rendered specular and diffuse lighting on the object. These regulaizations ae kept as they are as part of MaterialFusion. For more details we refer the reader to the nvdiffrecmc paper~\cite{hasselgren2022nvdiffrecmc}. We show relighting comparisons for the rest of the objects used in our evaluation in Figs.~\ref{fig:3d_comparison_3}--~\ref{fig:3d_comparison_5} along with the environment illumination used for rendering the training data images.

\paragraph{TensoIR Details.}

Given that TensoIR doesn't model metallicness, we added it as an additional parameter to the appearance representation decoded from the appearance tensor $\mathcal{G}_a$. The appearance features are interpolated from $\mathcal{G}_a$ and are then decoded with a radiance network $\mathcal{D}_c$, shading normal network $\mathcal{D}_\mathbf{n}$, and a material network $\mathcal{D}_\beta$ to produce the corresponding representations. Of interest is the material decoder which we modify to decode the additional metallicness parameter in addition to the albedo and roughness. All losses and hyperparameters are kept similar. Physically-based rendering is then used to render the resulting image, where metallicness is applied to the diffuse component of the Cook-Torrance reflectance model~\cite{cook_torrance}:

\begin{equation}
    f_r = (1 - m)\frac{\mathbf{a}}{\pi} + f_s
\end{equation}
where $f_s$ is the specular component of the Cook-Torrance model.

\begin{figure*}
    \centering
    \includegraphics[trim={0 1.0cm 0 0},clip,width=\textwidth]{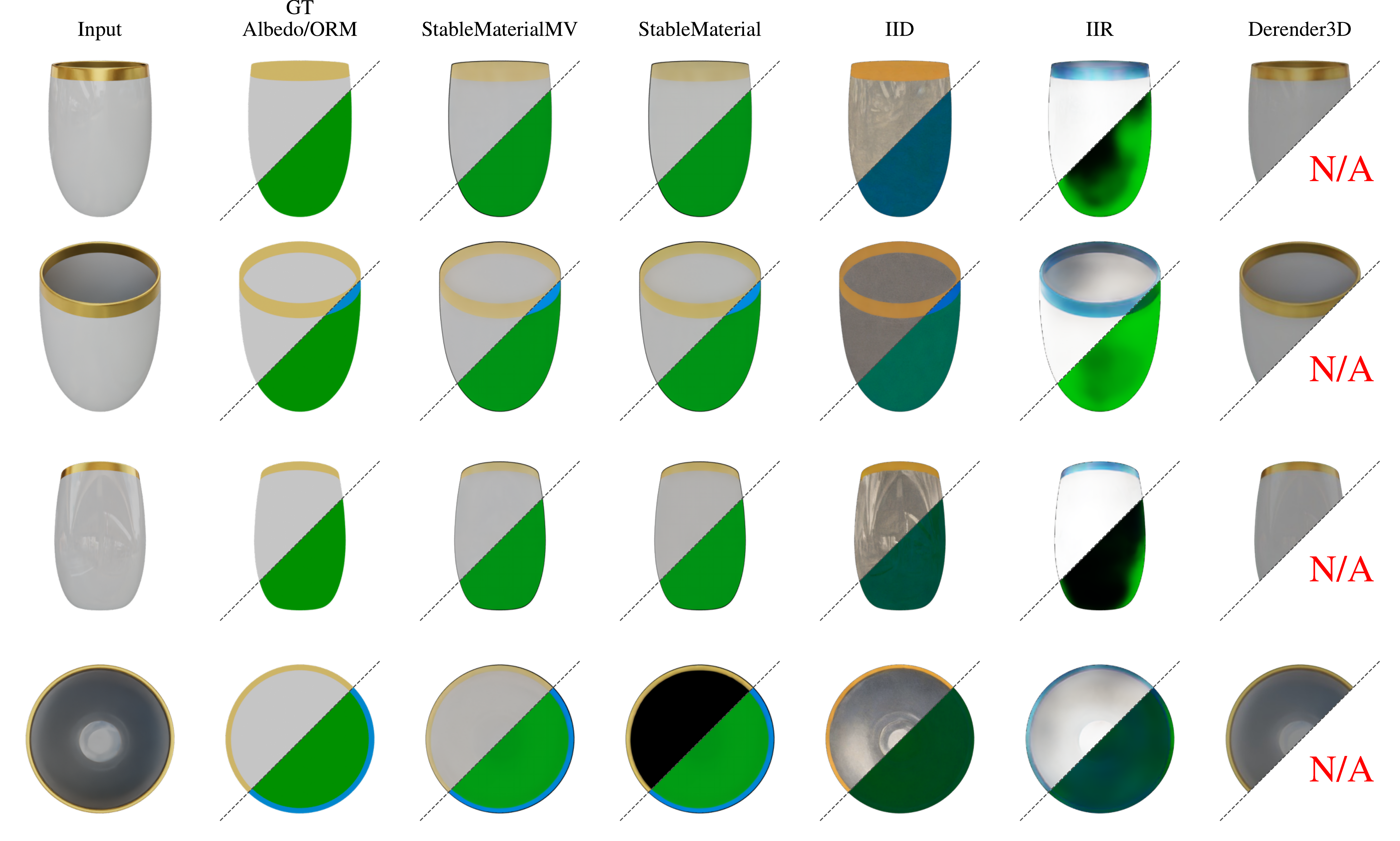}
    \label{fig:2d_comparison_1}
\end{figure*}
\begin{figure*}
    \centering
    \includegraphics[trim={0 1.0cm 0 0},clip,width=\textwidth]{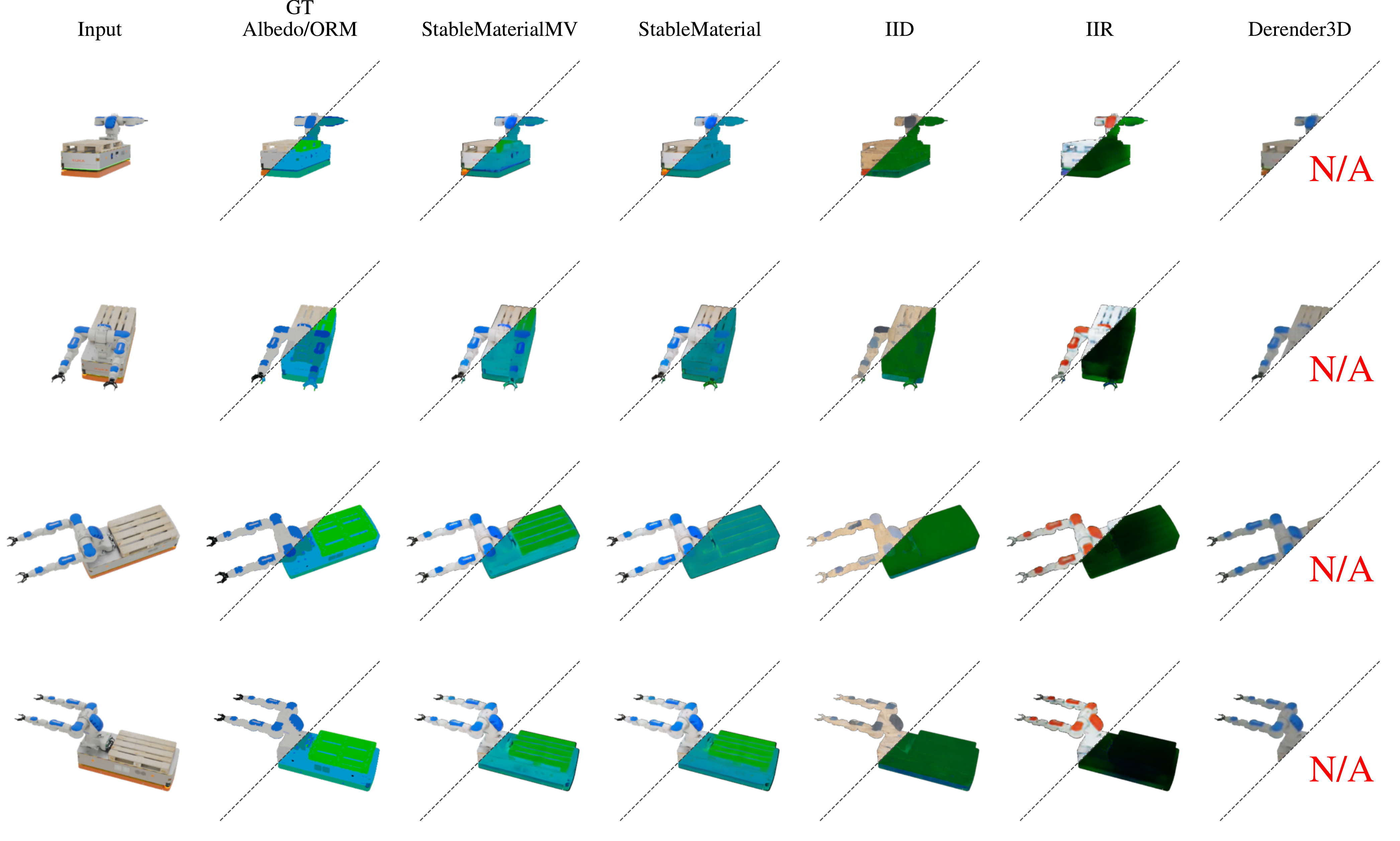}
    \caption{Full albedo and ORM comparison results for StableMaterial on the cup and armatures examples shown in Fig.~\ref{fig:2d_comparison}.}
    \label{fig:2d_comparison_2}
\end{figure*}
\begin{figure*}
    \centering
    \includegraphics[trim={0 1.0cm 0 0},clip,width=\textwidth]{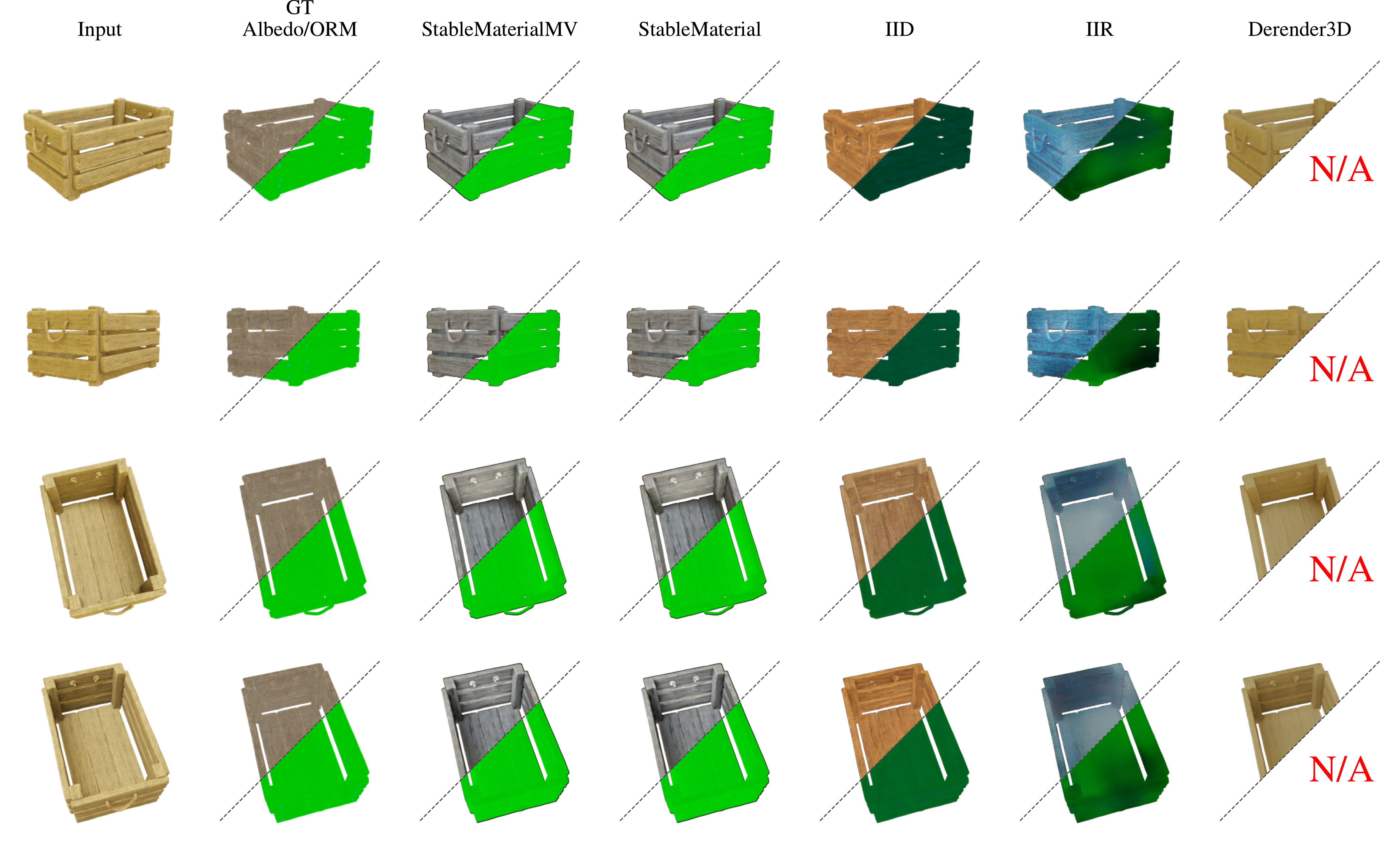}
    \label{fig:2d_comparison_3}
\end{figure*}
\begin{figure*}
    \centering
    \includegraphics[trim={0 1.0cm 0 0},clip,width=\textwidth]{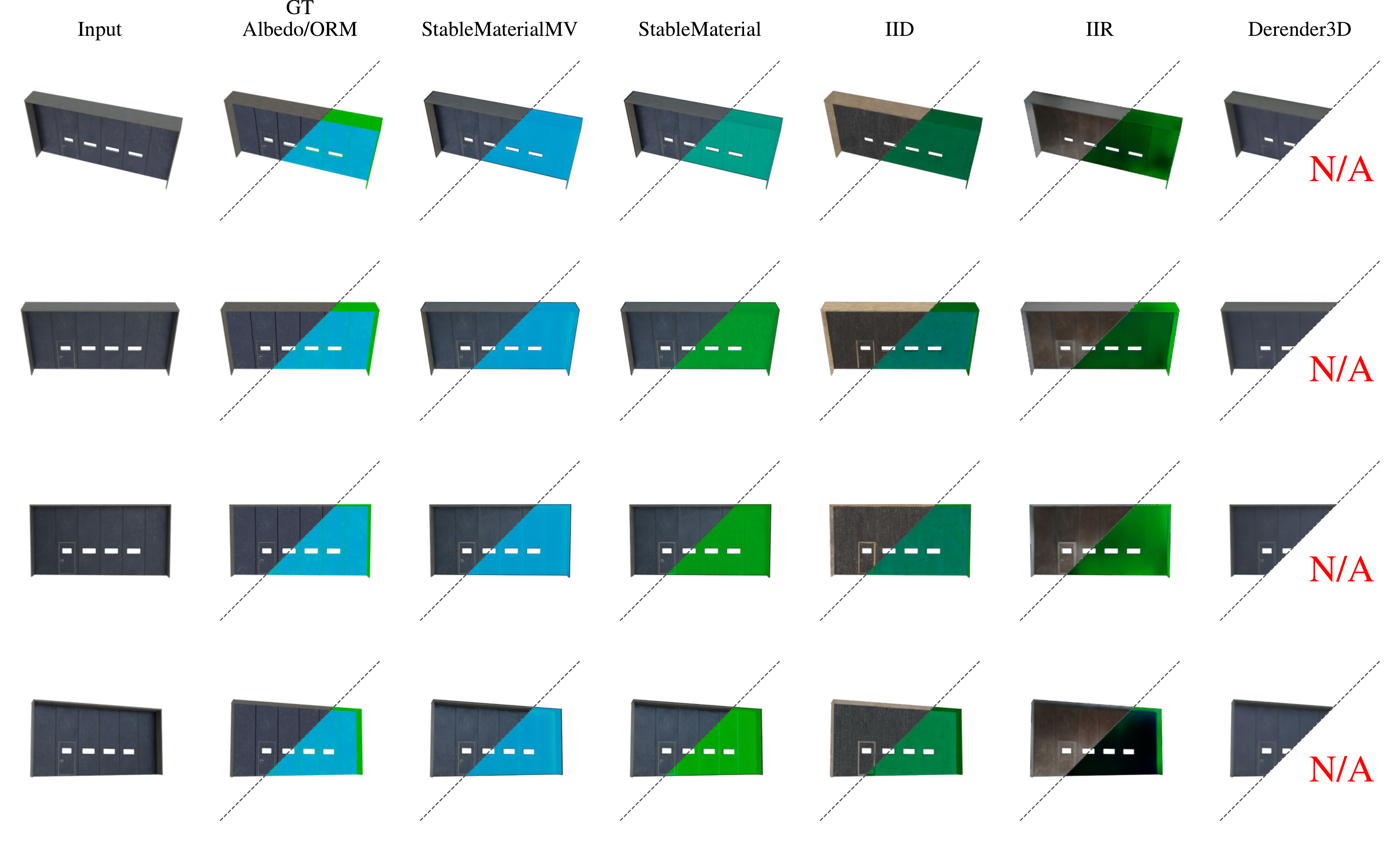}
    \caption{Additional albedo and ORM comparisons for randomly selected examples from the BlenderVault test dataset.}
    \label{fig:2d_comparison_4}
\end{figure*}
\begin{figure*}
    \centering
    \includegraphics[trim={0 1.0cm 0 0},clip,width=\textwidth]{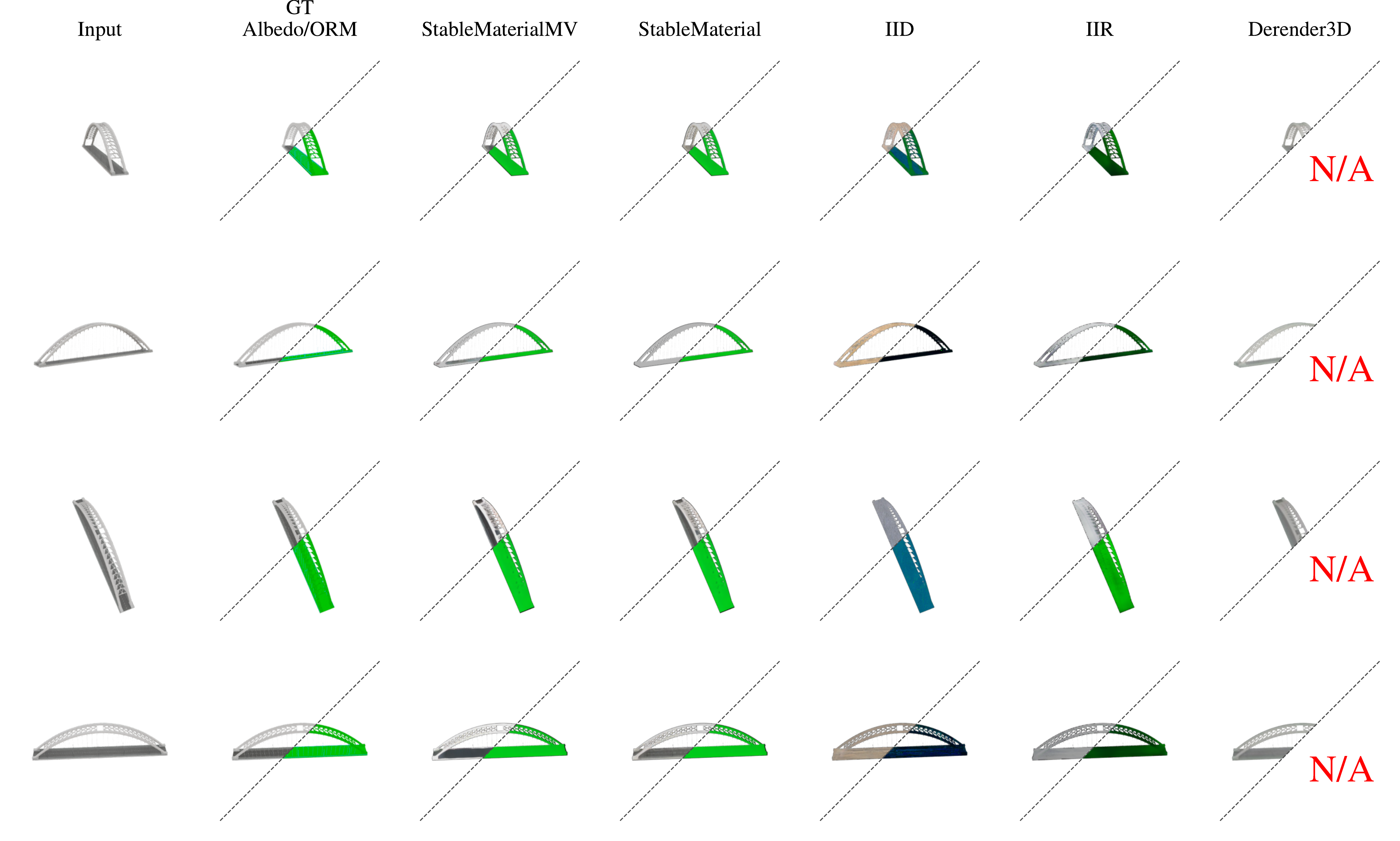}
    \label{fig:2d_comparison_5}
\end{figure*}
\begin{figure*}
    \centering
    \includegraphics[trim={0 1.0cm 0 0},clip,width=\textwidth]{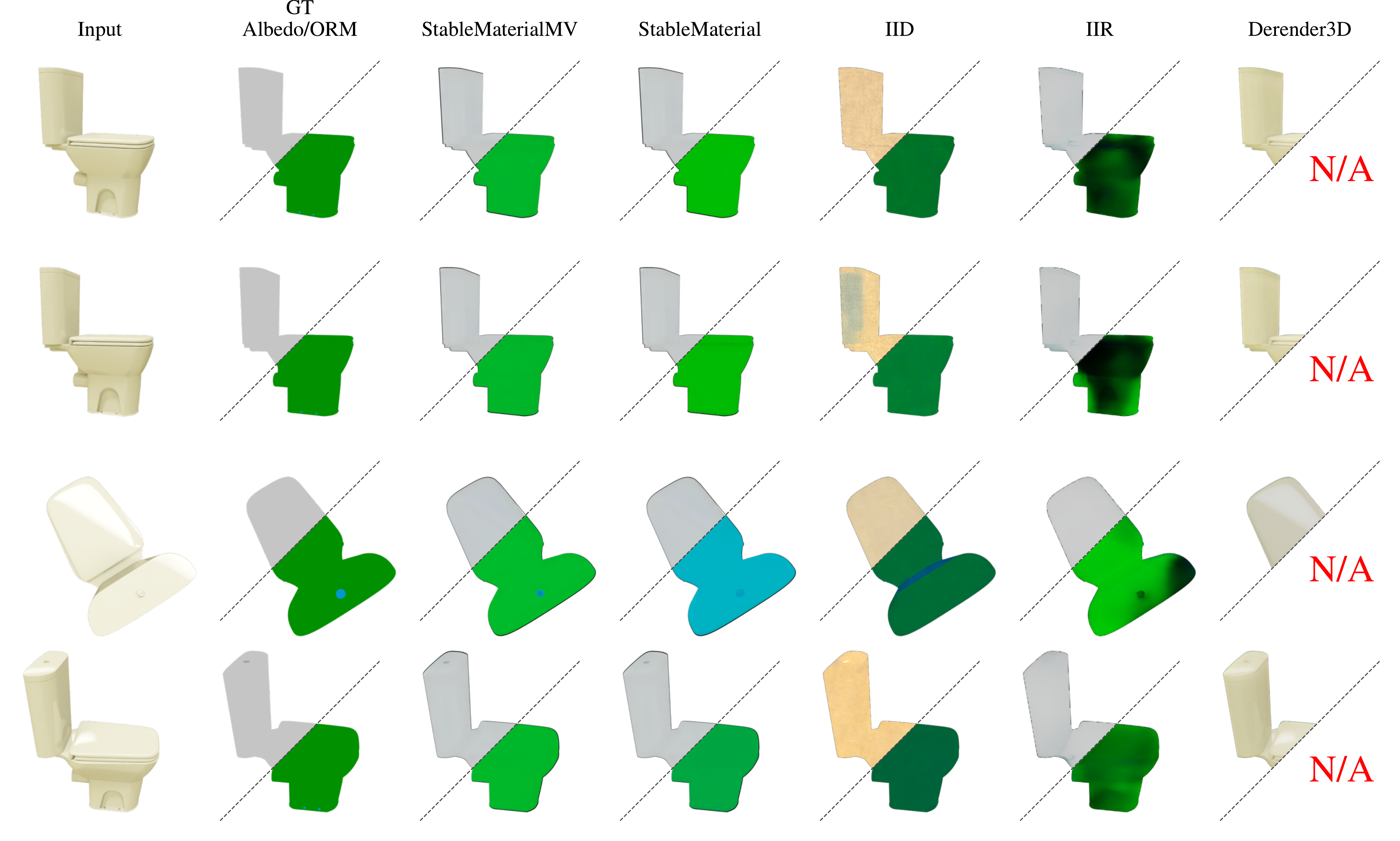}
    \caption{Additional albedo and ORM comparisons for randomly selected examples from the BlenderVault test dataset.}
    \label{fig:2d_comparison_6}
\end{figure*}
\begin{figure*}
    \centering
    \includegraphics[trim={0 0.5cm 0 0},clip,width=0.9\textwidth]{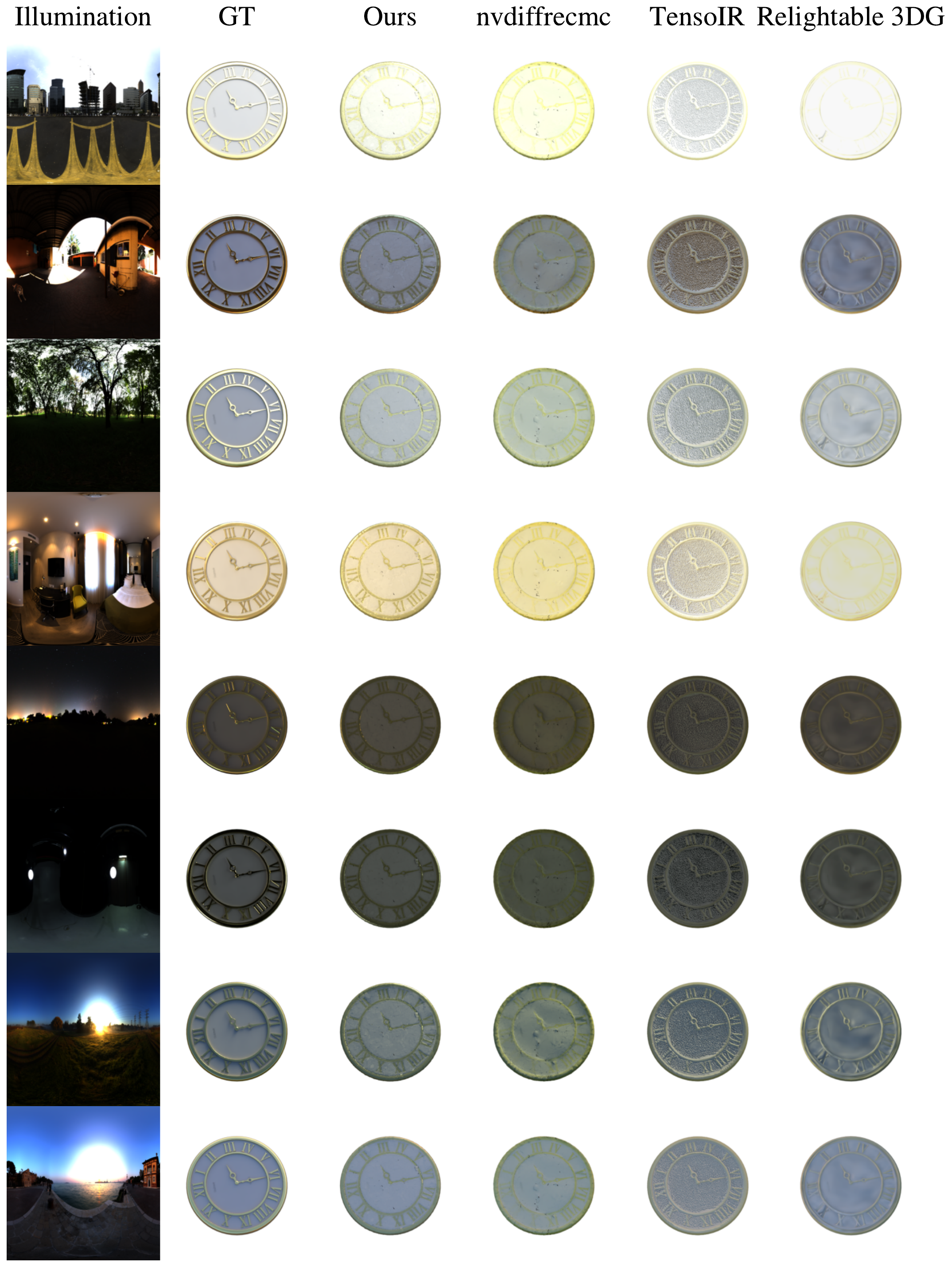}
    \caption{Comparison of MaterialFusion vs. other methods for relighting the clock object from the BlenderVault test dataset.}
    \label{fig:3d_comparison_clock}
\end{figure*}
\newpage
\clearpage

\end{document}